\documentclass[10pt, a4paper]{article}

\usepackage[final]{lrec2026} 

\usepackage[T1]{fontenc}
\usepackage[english]{babel}
\usepackage[dvipsnames]{xcolor}
\usepackage{hyperref}
\usepackage{url}
\usepackage[capitalise,noabbrev]{cleveref}
\usepackage{tabularray}
\UseTblrLibrary{booktabs}
\usepackage{enumitem}
\usepackage{tikz}
\usepackage{relsize}

\usepackage{caption}
\usepackage{subcaption}

\newcommand{\SB}{\textsc{SemBench}}
\newcommand{\SBdef}{\SB\textsubscript{Def}}
\newcommand{\SBex}{\SB\textsubscript{Ex}}

\newcommand{\wic}{\textsc{WiC}}

\newcommand*\circled[2][]{\tikz[baseline={(char.base)}]{\node[shape=circle,draw,font=\relsize{-1},inner sep=1pt,#1] (char) {#2};}}

\title{\SB: A Universal Semantic Framework for LLM Evaluation}

\name{\shortstack{Mikel Zubillaga, Naiara Perez, Oscar Sainz, German Rigau}}
        \address{HiTZ Center - Ixa, University of the Basque Country UPV/EHU \\
         \{\textit{name.surname}\}@ehu.eus\\}

\abstract{
Recent progress in Natural Language Processing (NLP) has been driven by the emergence of Large Language Models (LLMs), which exhibit remarkable generative and reasoning capabilities. However, despite their success, evaluating the true semantic understanding of these models remains a persistent challenge. Traditional benchmarks such as Word-in-Context (\wic{}) effectively probe this capability, but their creation is resource-intensive and often limited to high-resource languages. In this paper, we introduce \SB{}, a framework for automatically generating synthetic benchmarks that assess the semantic competence of LLMs using only dictionary sense definitions and a sentence encoder. This approach eliminates the need for curated example sentences, making it both scalable and language-independent. We evaluate \SB{} in three languages (English, Spanish, and Basque) spanning different levels of linguistic resources, and across a wide range of LLMs. Our results show that rankings derived from \SB{} strongly correlate with those obtained from standard \wic{} datasets. Furthermore, our analysis demonstrates that only a small number of examples is required to achieve stable and meaningful rankings. Overall, \SB{} provides a lightweight, adaptable, and data-efficient framework for cross-lingual evaluation of semantic understanding in LLMs.
 \\ \newline  }

\begin{document}

\maketitleabstract

\section{Introduction}
\label{sec:introduction}

In recent years, the field of Natural Language Processing (NLP) has experienced significant advancements, largely driven by the development of Large Language Models (LLMs). Trained on massive datasets, these models have demonstrated impressive capabilities in generating coherent, human-like text. Today, LLMs are not only employed for traditional tasks such as summarization and translation, but are also increasingly used as autonomous agents, programming assistants, and even literature reviewers~\citep{jiang-etal-2025-towards,dong2025surveycodegenerationllmbased,liao2024llmsresearchtoolslarge}. This shift in how LLMs are applied has also prompted a reevaluation of how they are assessed—moving away from static benchmarks toward more dynamic, context-sensitive evaluations, often inspired by methodologies from reinforcement learning. However, most of the traditional NLP tasks remain unsolved, even for the best performing state-of-the-art approaches.

One such evaluation approach is the Word-in-Context (\wic{}) challenge \citep{pilehvar2019wic}. As the name suggests, \wic{} assesses a model's ability to distinguish between different senses of the same word based on context. Specifically, the task presents a target word used in two separate sentences and asks the model to determine whether the word carries the same meaning in both instances or reflects different senses. While this may appear straightforward--especially for \textit{language} models--it has proven to be quite challenging, with performance often only slightly better than random guessing~\citep{hayashi-2025-evaluating}.

Developing a \wic{} dataset for a given language can be as simple as extracting sense-specific examples from a dictionary. However, many dictionaries either lack usage examples altogether or are restricted by licensing constraints. Alternatively, manually constructing a \wic{} dataset is resource-intensive, requiring significant effort from linguistic experts~\cite{goworek-etal-2025-senwich}. 

In this work, we introduce \textbf{\SB{}}, a novel fully automatic framework for evaluating the semantic competence of LLMs. Rather than relying on preconstructed datasets, \SB{} performs evaluation through generation, using only a dictionary with sense definitions (a resource typically more accessible than dictionaries containing usage examples) and a sentence encoder. This design makes \SB{} both scalable and language-independent, enabling consistent evaluation even in low-resource settings. We apply our methodology across three typologically diverse languages (English, Spanish, and Basque) and a broad range of LLM families and sizes. Experimental results show that the model rankings by \SB{} strongly correlate with those obtained from standard \wic{} datasets, validating its effectiveness. Moreover, ablation studies confirm the efficacy and practicality of \SB{}, as only a small number of instances are required to achieve stable and meaningful results.

In summary, our main contributions are as follows: \textbf{(1)} we present \SB, a fully automatic methodology for evaluating semantic understanding in LLMs through text generation, which yields results strongly aligned with \wic{}; \textbf{(2)} we demonstrate the adaptability of \SB{} across languages with varying resources levels---high (English), moderate (Spanish), and low (Basque); \textbf{(3)} we analyze the impact of the number of test instances and benchmark size, showing that minimal data are sufficient to produce stable and interpretable rankings; and, \textbf{(4)} we propose a simple yet effective heuristic for controlling evaluation difficulty, which accurately reflects task complexity while preserving high correlation with \wic{} performance.

\begin{figure*}
    \centering
\includegraphics[width=\linewidth,trim={20pt 5pt 20pt 10pt},clip]{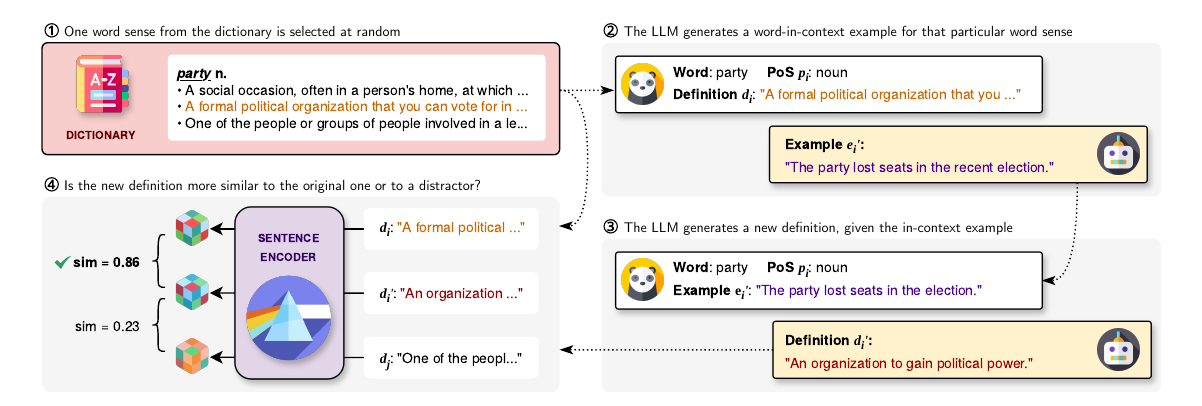}
    \caption{A general overview of the \SB{} framework.}
    \label{fig:sembench}
\end{figure*}

\section{Related Work}
\label{sec:related-work}

\subsection{Evaluating the semantic capabilities on LLMs}
Although human evaluation remains the most reliable method for assessing the quality of generated responses, particularly since the emergence of LLMs, it is both time-consuming and difficult to scale when comparing multiple models or variations. With the rapid progress in the field, the need for efficient automated evaluation methods has become increasingly evident.

One of the most influential automatic benchmarks before the emergence of LLMs was SuperGLUE \cite{wang2020supergluestickierbenchmarkgeneralpurpose}, a language understanding benchmark that extends the original GLUE benchmark \cite{wang-etal-2018-glue}. SuperGLUE introduced a set of more challenging sub-tasks designed to test various aspects of language model performance. Among these tasks is Word in Context \cite[\wic;][]{pilehvar2019wic}, which focuses on evaluating the semantic capabilities for a given language model. 
More precisely, \wic{} presents a scenario in which the same word appears in two different contexts. The model is then asked to determine whether the word has the same meaning in both sentences or, conversely, whether each occurrence conveys a different sense. Although \wic{} was originally designed to evaluate context-sensitive word embeddings rather than LLMs, recent work has shown that it also serves as an effective benchmark for assessing LLM semantic understanding \citep{hayashi-2025-evaluating}.
In this work, we leveraged \wic{} to generate the model rankings that we will consider as gold standard for our methodology, \SB{}. 

\subsection{Sense representations in LLMs}
Transformer architectures have fundamentally changed representation learning by producing context-sensitive embeddings for tokens rather than single, fixed vectors. Whereas static embeddings assign one position per word type \citep{pennington-etal-2014-glove}, Transformer-based models produce a family of context-dependent vectors, which change according to the context. This richer representational palette has practical value for the study of lexical semantics \citep{petersen-potts-2023-lexical}: geometric relationships among contextual embeddings offer a powerful means of distinguishing nuances in word meaning.

With the advent of recently developed LLMs, automatically generated sense definitions provide a new avenue for representing word usage~\cite{periti-etal-2024-automatically}. These definitions effectively capture the underlying meaning of a word, even in cases of polysemy, making them a valuable and interpretable tool for word usage representation \citep{gardner2022definition}. In this work, we aim to indirectly measure the semantic capabilities of LLMs by measuring the quality of the generated definitions.

\subsection{Generating Word Sense Definitions and Examples}
Early works on generating word sense definitions were motivated by the goal of improving the interpretability of static embeddings \citep{gadetsky-etal-2018-conditional, mickus-etal-2022-semeval}. The original formulation enhanced models to generate a natural-language definition from a single static embedding of the target word \citep{Noraset_Liang_Birnbaum_Downey_2017}. Due to the polysemous nature of some words, producing accurate definitions from the word embedding alone proved difficult; consequently, the paradigm shifted: instead of relying solely on the word, contextual information was incorporated to generate more appropriate and precise definitions \citep{ishiwatari-etal-2019-learning, huang-etal-2021-definition, zhang-etal-2022-fine-grained}. To do so, recent works have started using LLMs to create those definitions \citep{giulianelli-etal-2023-interpretable, periti-etal-2024-automatically}.

In addition to definitions, context or example sentences where the word is used can also be generated. This is a key challenge in understanding and modeling language semantics. Some approaches rely on training models with existing dictionaries or corpora, pairing headwords with illustrative sentences to automatically create new examples \citep{barba2021exemplification, he-yiu-2022-controllable}. In contrast, \citet{harvill-etal-2023-one} introduced a more flexible method, showing that meaningful sentences can also be produced using just a single reference sentence as input. More recently, \citet{cai-etal-2024-low} show that example generation can be done using LLMs in a zero-shot setting, only using as input the definition of the headword. However, previous works focus on definition generation without the purpose of evaluating model's semantic capabilities. In this work, we focus on this topic and present \SB{}, an automatic framework for semantic capabilities evaluation.

\section{\SB}
\label{sec:sembench}

\SB{} is a novel, fully automatic framework for evaluating the semantic understanding capabilities of LLMs. The approach rests on the intuition that a model demonstrating genuine semantic competence should be able to transition consistently between \emph{definitions} and \emph{usage examples} corresponding to the same sense of a word. Unlike previous evaluation protocols that rely on curated datasets or manual annotation, \SB{} constructs test instances directly from existing lexical resources. Its overall workflow is summarized in~\cref{fig:sembench}, which illustrates how words, senses, and model generations interact within the framework. The remainder of this section details the core resources and methodological components of \SB{}.

\subsection{Resources}
\label{ssec:sembench-resources}

\SB{} relies on two key components: a \textbf{sentence encoder} that is used to compute the semantic similarity of definition pairs, and a \textbf{dictionary}.
Formally, let the dictionary be represented as

\[ D = \{(w, S(w)) \;|\; w \in W\}, \]

\noindent where $W$ is the set of all words contained in the dictionary, and each word $w \in W$ is associated with a finite set of senses:

\[ S(w) = \{s_1, s_2, \dots , s_{n_w}\},\; n_w \geq 1. \]

\noindent Each sense $s_i \in S(w)$ is represented as a triplet $s_i = (d_i, p_i, e_i)$, where $d_i$ denotes the textual definition of the sense, $p_i$ is the part-of-speech (PoS) label, and $e_i$ is an optional usage example. That is, every word is associated with one or more senses, each defined by at least a definition and grammatical category and, possibly, accompanied by an in-context example.

\subsection{Methodology}
\label{ssec:sembench-methodology}

\SB{} begins by sampling a polysemous word $w \in W$, such that $n_w > 1$.
From its set of senses $S(w)$, one particular sense $s_i$ is chosen at random and used as the seed for generation. Depending on the available dictionary format, \SB{} provides two experimental configurations:
\begin{itemize}
    \item \textbf{From definitions} (\SBdef): the default setup, shown in~\cref{fig:sembench}, which does not assume access to in-context examples. The LLM is asked to \textit{i)} generate a usage example $e_i'$ for $w$ given its definition $d_i$ and PoS $p_i$; and \textit{ii)} generate a dictionary definition $d_i'$ for word $w$ given its PoS $p_i$ and the synthetic example $e_i'$.
    \item \textbf{From examples} (\SBex): a simpler setup that assumes the dictionary provides a usage example $e_i$. The LLM is asked to generate a dictionary definition $d_i'$ for word $w$ given its PoS $p_i$ and the context $e_i$ (that is, it bypasses step \circled{2} in~\cref{fig:sembench}).
\end{itemize}

\noindent The synthetic definition $d_i'$ is then compared against two reference definitions from the same dictionary entry: the \textbf{target definition} $d_i$, which corresponds to the intended sense, and a \textbf{distractor definition} $d_j$, associated with a different sense $s_j$ of the same word $w$ ($i \neq j$). The model is considered correct if its definition is more semantically similar to the target than the distractor, according to an encoder-based similarity metric:

\[ sim(d_i', d_i) >  sim(d_i', d_j), \]

\noindent where similarity is computed as the dot product of the corresponding embedding representations. Model performance is then quantified as the proportion of correctly identified senses over a set of $N$ randomly selected test instances ($d_i$, $d_j$).

\begin{table*}
    \centering
    \begin{tblr}{
        colspec={llXX},
        cells={font=\scriptsize},
        row{1}={c,font=\scriptsize\bfseries},
        rowsep=1pt
    }
    \toprule
        Language & Reference \wic{} & Dictionary & Sentence Encoder \\
    \midrule
        (EN) English  & \citet{pilehvar2019wic} & Oxford Dictionary of English (ODE)\footnotemark[1] & \SetCell[c=1,r=3]{c,m,bg=gray9} {EmbeddingGemma 300M\\\cite{vera2025embedding}} \\
        (ES) Spanish  & \citet{PLN6683} & \textit{Diccionario de la RAE} (DRAE)\footnotemark[2] \\
        (EU) Basque   & \citet{urbizu-etal-2022-basqueglue} & \textit{Egungo Euskararen Hiztegia} (EEH)\footnotemark[3] \\
        \bottomrule
    \end{tblr}
    \caption{Resources employed in the experimentation for each language.}
    \label{tab:experimental-setting}
\end{table*}

\section{Experimental Setup}
\label{sec:experimental-setup}

To validate the effectiveness of \SB, we compare its results against a well-established word sense disambiguation evaluation framework: Word-in-Context (\wic{}). Specifically, we evaluate a set of LLMs with both methodologies and measure the degree of correlation between the two frameworks using the \textbf{Spearman's rank correlation coefficient} ($\rho$). A high correlation would indicate that \SB{} captures a notion of semantic understanding comparable to traditional sense discrimination tasks, while requiring no manual annotation. 

Moreover, we conduct experiments in three typologically diverse languages with varying levels of resource coverage, each of which has an existing \wic{} benchmark for reference: \textbf{English} (Germanic/high), \textbf{Spanish} (Romance/moderate), and \textbf{Basque} (isolate/low).

Below, we first detail the \wic{} datasets employed to establish reference rankings, then introduce the resources used to build and run \SB{} in each language, and finally outline the evaluated LLMs and their inference setups.

\footnotetext[1]{\url{www.oed.com}; through~\citet{TFGLeireValera}.}
\footnotetext[2]{\url{www.rae.es/}; accessed via the public website.}
\footnotetext[3]{\url{www.ehu.eus/eeh}; SQL dump \texttt{2024-01-26}.}
\addtocounter{footnote}{3}

\subsection{Computation of Reference Rankings}

\cref{tab:experimental-setting} reports the existing \wic{} datasets that we rely on to establish a reference ranking of models per language. These datasets contain pairs of sentences featuring the same target word $w$ used in two distinct contexts. Each pair is annotated with a binary label indicating whether $w$ conveys the \emph{same sense} in both contexts or not. To evaluate each LLM on this task, we follow a configuration similar to \SBex{}: for each \wic{} instance, the model is presented with the two contexts $e_i$ and $e_j$ in which $w$ appears, and is asked to generate a dictionary-style definition for each occurrence, denoted $d_i'$ and $d_j'$. Then, we encode both $d_i'$ and $d_j'$ using the same encoder employed throughout the experiments (introduced next), and compute their cosine similarity. The model is considered to predict \emph{same sense} if $sim(d_i', d_j') > 0.5$,\footnote{Although the range of the similarity metric is defined as $[-1, 1]$, we considered $0.5$ as the default decision threshold because $0$ is interpreted as not similar and $1$ is interpreted as perfectly similar.} and \emph{different sense} otherwise. Accuracy is then calculated as the proportion of correctly classified \wic{} pair instances according to the gold labels.

\subsection{Experimental Resources}
\label{ssec:experiment-resources}

\begin{table}
    \begin{tblr}{
            width=\linewidth,
            colspec={X|rlrlrl},
            columns={font=\scriptsize},
            column{3,5,7}={font=\tiny,leftsep=2pt},
            column{2,4,6}={rightsep=0pt},
            row{1}={font=\scriptsize\bfseries},
            rowsep=0.65pt
        }
        \toprule
        Statistic & \SetCell[c=2]{c} English && \SetCell[c=2]{c} Spanish && \SetCell[c=2]{c} Basque \\
        \midrule
        Sense density     &   6.12 & $\pm$ 3.5  &  6.92 & $\pm$ 5.9  &  2.07 & $\pm$ 0.3  \\
        Definition length &  64.77 & $\pm$ 27.7 & 59.77 & $\pm$ 31.7 & 39.99 & $\pm$ 28.1 \\
        Example length    & 106.10 & $\pm$ 42.5 & \SetCell[c=2]{c} n/a && \SetCell[c=2]{c} n/a \\
        \bottomrule
    \end{tblr}
    \caption{Statistics of \SB{}\textsubscript{ \textsc{rand}}. Lengths are reported in terms of number of characters.}
    \label{tab:sembench_statistics}
\end{table}

For each language, we created a \textbf{test set of 1,000 instances} sampled at random from the corresponding lexical resource (see~\cref{tab:experimental-setting}; our code and processed resources are available online).\footnote{\url{https://github.com/MikelZubi/SemBench}} Each instance consists of a target definition and a distractor definition belonging to a different sense of the same word, selected according to the procedure described in~\cref{ssec:sembench-methodology}. 

\paragraph{\SBdef{} vs \SBex{}.} As shown in \cref{tab:sembench_statistics}, only the English subset includes example sentences alongside sense definitions. Consequently, the comparisons between \SBdef{} and \SBex{} were conducted exclusively for English. To ensure a fair comparison, we used the same set of senses across both evaluation variants. The Spanish and Basque subsets were evaluated using \SBdef{} only.

\paragraph{Difficulty levels.}
To better characterize the evaluation space, we constructed four sub-datasets of varying difficulty by controlling the semantic similarity between the target and distractor definitions. For each word, all alternative definitions were first ranked by their cosine similarity to the target definition, computed using the sentence encoder. Then, the distractor was selected according to one of four strategies:
\begin{itemize}[noitemsep]
    \item \textsc{easy}: the least similar definition.
    \item \textsc{mid}: a definition from the middle of the list.
    \item \textsc{hard}: the most similar definition.
    \item \textsc{rand}: a randomly selected definition.
\end{itemize}
For each difficulty level, we sampled 1,000 pairs of senses from the English dictionary, in order to compare its effect on the \SBdef{} and \SBex{} strategies.

\paragraph{Sentence encoder.} As for the sentence encoder, we chose the EmbeddingGemma model of 300M parameters~\cite{vera2025embedding} for all the experiments, given its multilingual support and demonstrated competitiveness.

\subsection{Models and Inference Details}
\label{ssec:experiment-models}

We evaluated a diverse set of open-weight LLMs, covering a range of architectures, sizes, and training paradigms. All models were accessed through their instruction-tuned or chat variants to ensure consistent prompting behavior. The evaluated models include:
\begin{itemize}
    \item \textbf{Gemma 3} \cite{gemmateam2025gemma3technicalreport}: Multimodal instructed decoder only models. We have used the 4B, 12B, and 27B Instruct variants.
    \item \textbf{Qwen3} \cite{yang2025qwen3technicalreport}: Reasoning instructed decoder only models. We have used the 4B, 8B, 14B, and 32B Instruct variants.
    \item \textbf{Llama 2} \cite{touvron2023llama}: Instructed decoder only models. We have used the 7B Instruct variant.
    \item \textbf{Llama 3.1} \cite{grattafiori2024llama3herdmodels}: Instructed decoder only models. We have used the 8B and 70B Instruct variants.
    \item \textbf{Latxa Instruct} \cite{sainz-etal-2025-instructing}: Instructed decoder only models specialized in the Basque language, built on Llama 3.1. We have used the 8B and 70B variants.
\end{itemize}

We ran experiments under both \textbf{zero-shot} and \textbf{5-shot} configurations, following the same prompting scheme across all languages and \SB{} variants \SBdef{} and \SBex. Prompts were designed to elicit definitions or examples as described in~\cref{ssec:sembench-methodology}; full templates are provided in Appendix \ref{app:prompts}. To ensure reproducibility, we generated all outputs using a greedy decoding strategy. Regarding the examples used for few-shot prompting, whenever possible we sampled definition–example pairs for each sense. However, since not all dictionaries include examples alongside definitions, we manually created the missing ones. As only five examples were needed, this did not require substantial manual effort. 

\section{Results}

To validate our hypothesis, we present the following experimental results. We begin by comparing the rankings produced by our proposed method with those obtained using \wic{}, our gold standard. Next, we extend the analysis to moderate- and low-resource language evaluations to demonstrate the applicability of \SB{}. We also investigate how the number of examples in \SB{} affects the final rankings, highlighting the robustness of our approach. Finally, we present the overall results obtained with \SB{}.

\subsection{Validation of \SB{} against \wic{}}
\label{ssec:validation}

Our analysis covers three languages representing different levels of linguistic resources, allowing us to assess the robustness and general applicability of the method. The following sections present and discuss the results obtained for each language.

\begin{figure}
    \centering
    \begin{subfigure}[b]{0.49\linewidth}
        \includegraphics[width=\linewidth]{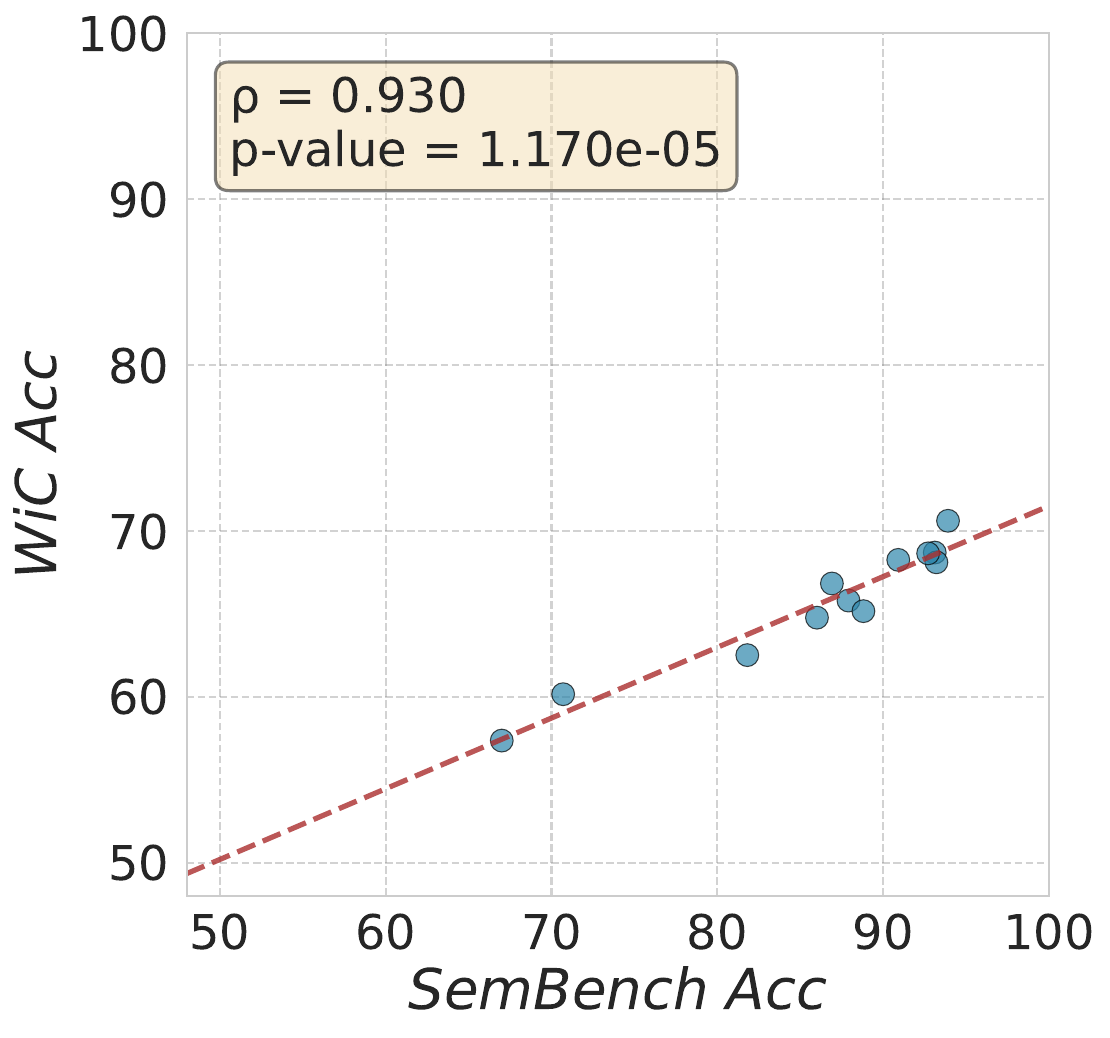}
        \caption{\SBdef{}}
        \label{fig:scatter_EN_F_DEF}
    \end{subfigure}
    \begin{subfigure}[b]{0.49\linewidth}
        \includegraphics[width=\linewidth]{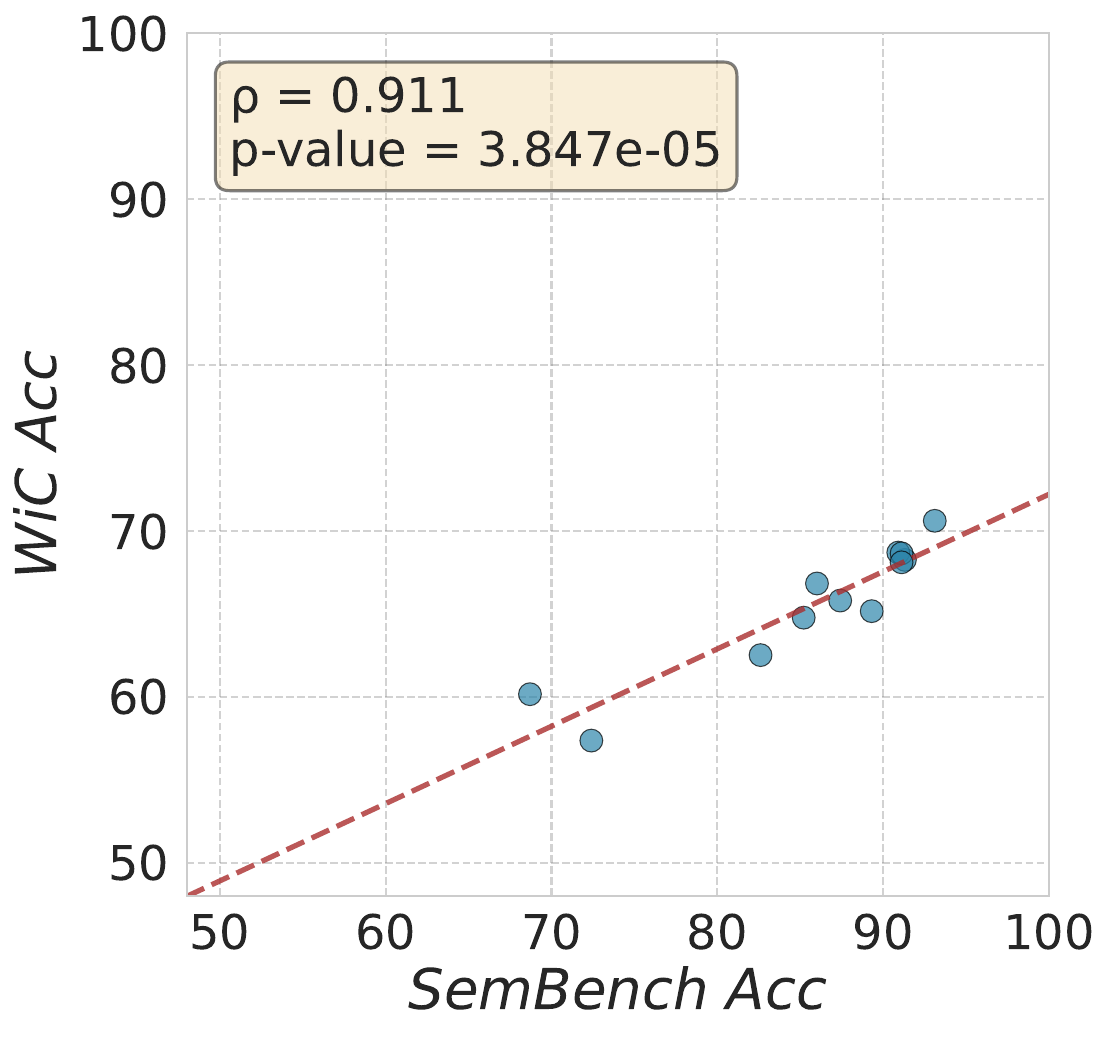}
        \caption{\SBex{}}
        \label{fig:scatter_EN_F_EXP}
    \end{subfigure}
    \caption{Accuracy (Acc) correlation between \SB{} and English \wic{} using 5-shot configuration. On the left, we compare \SB{\textsubscript{\textsc{}rand}} starting from definitions, whereas on the right, we compare the variant starting from examples.}
    \label{fig:scatter_EN}
\end{figure}

\begin{figure}
    \centering
    \begin{subfigure}[b]{0.49\linewidth}
        \includegraphics[width=\linewidth]{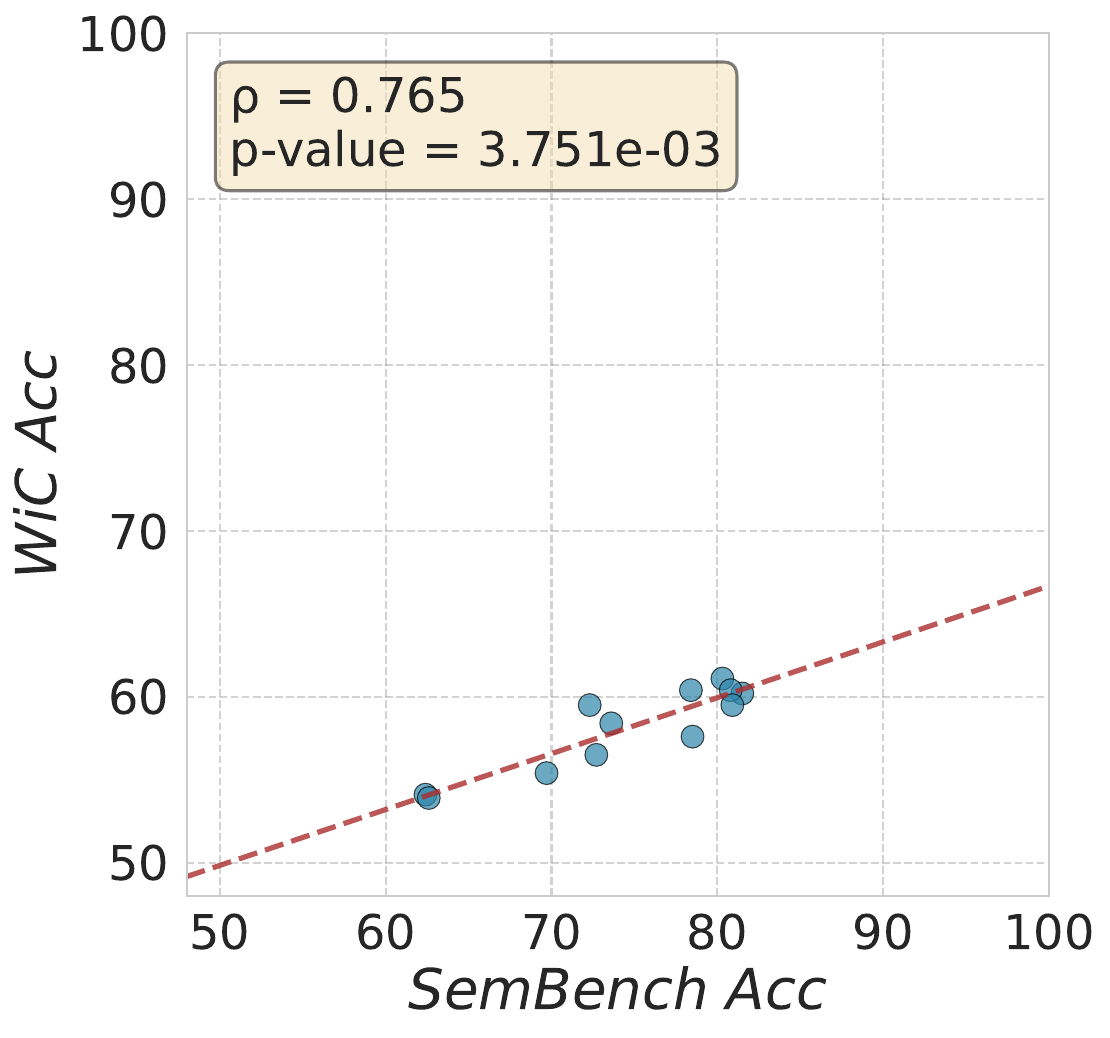}
        \caption{Spanish}
        \label{fig:scatter_ES}
    \end{subfigure}
    \begin{subfigure}[b]{0.49\linewidth}
        \includegraphics[width=\linewidth]{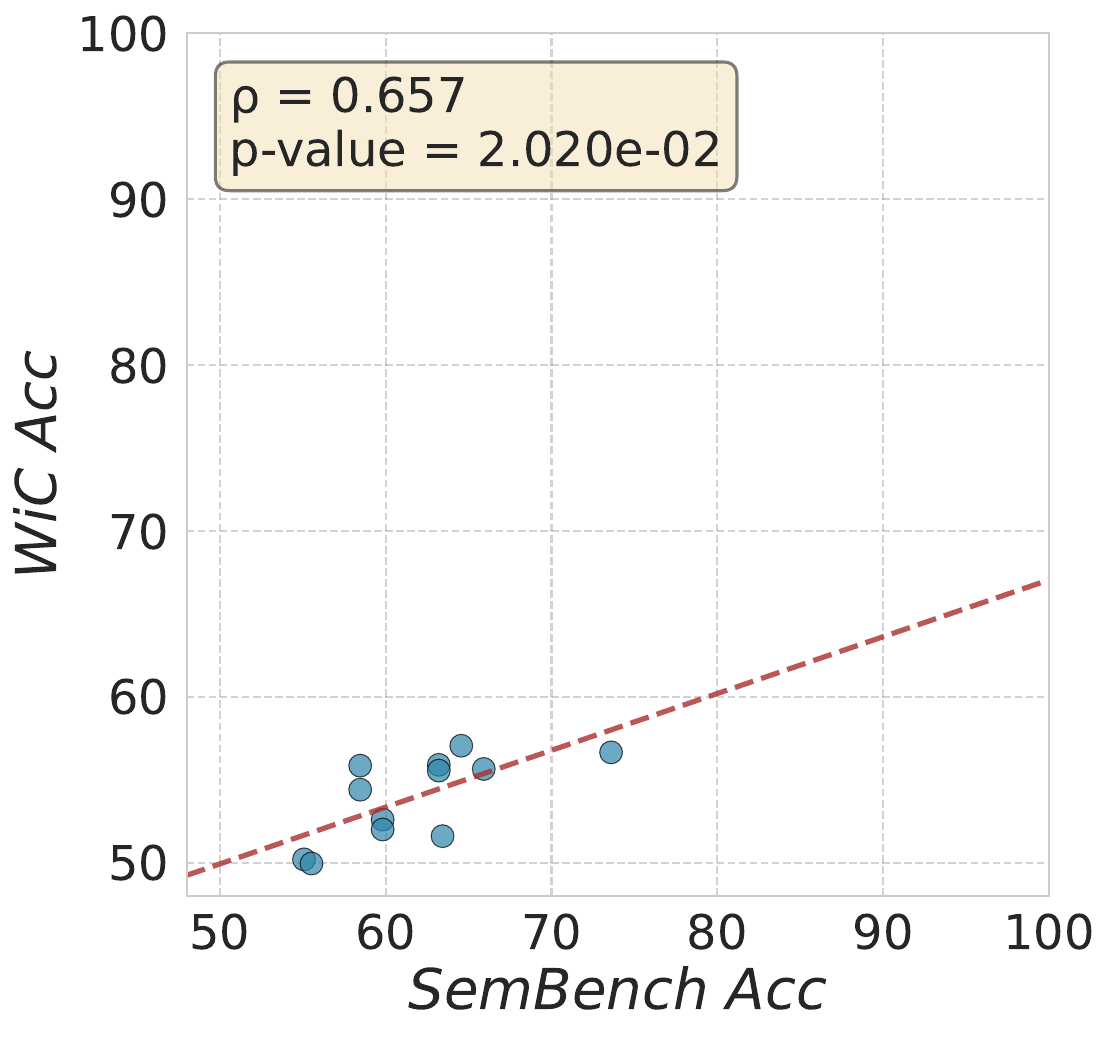}
        \caption{Basque}
        \label{fig:scatter_EU}
    \end{subfigure}
    \caption{Accuracy (Acc) correlation between \SBdef{} with Spanish (left) and Basque (right) \wic{} datasets on the 5-shot settings.}
    \label{fig:scatter_EUES}
\end{figure}

\begin{figure*}
    \centering
    \includegraphics[width=\linewidth]{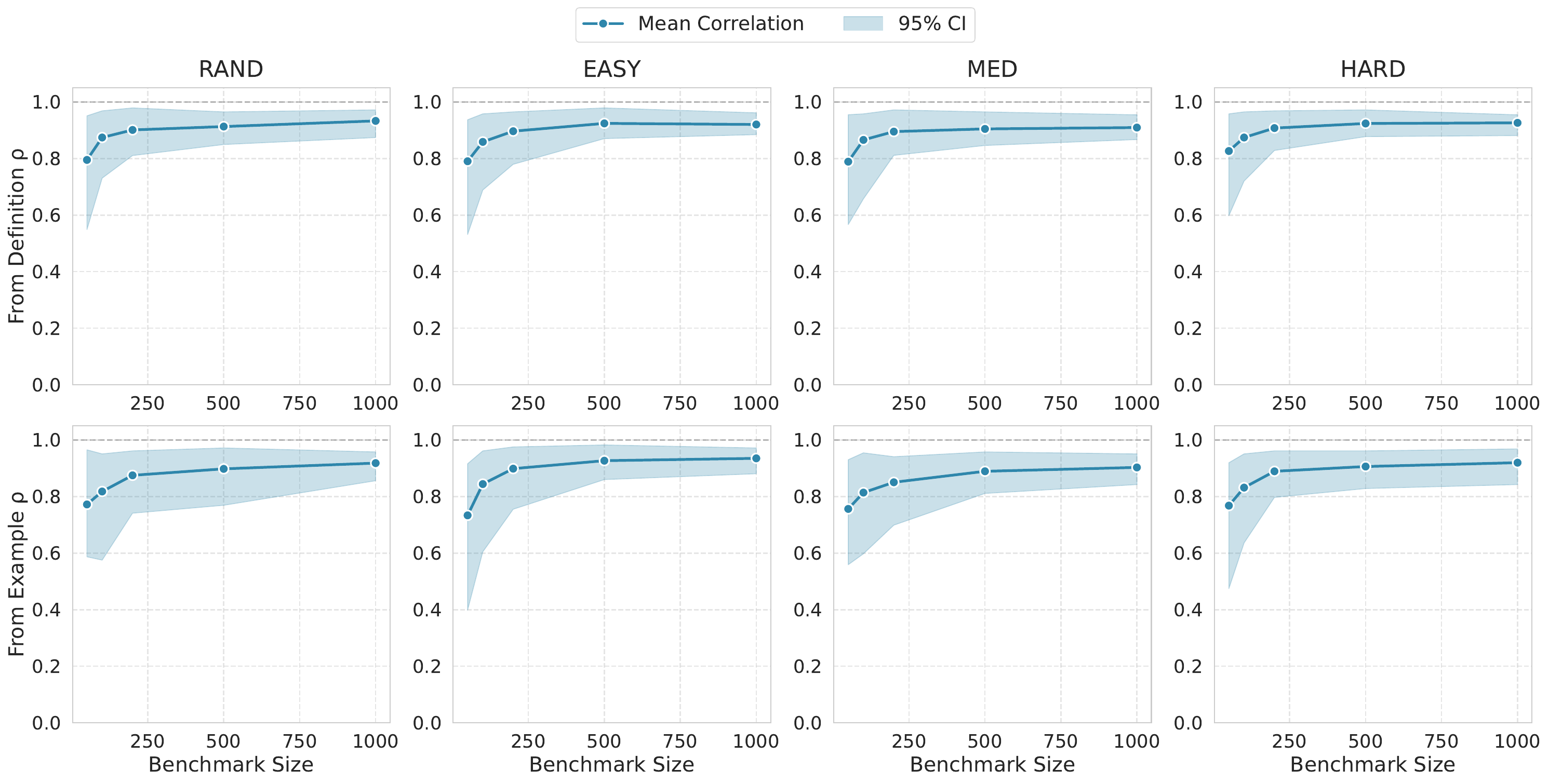}
    \caption{Spearman's $\rho$ correlation curves between different \SB{} variants and \wic{} with varying data-points. Confidence intervals are obtained by applying bootstrapping through 100 iterations.}
    \label{fig:boostraping_stability}
\end{figure*}

\paragraph{Validation against English \wic{}.} \cref{fig:scatter_EN} illustrates the correlation between the scores obtained by the two variants of our proposed method (namely, \SBdef{} and \SBex{}) and those from \wic{} for English. All the scores are obtained using 5 few-shot examples. We observe that, although our benchmarks tend to yield slightly higher absolute scores (up to 95 vs 70 points), they exhibit an almost perfect correlation with \wic{} ($\rho=0.930$ and $\rho=0.911$), demonstrating the strong reliability and validity of our approach. Moreover, the range of values across models is noticeably broader, which suggest that \SB{} offers greater discriminative capacity; in contrast, \wic{} results tend to cluster more tightly, particularly among high-performing models. A closer examination reveals only minor differences between the two variants, \SBdef{} (\cref{fig:scatter_EN_F_DEF}) and \SBex{} (\cref{fig:scatter_EN_F_EXP}). This observation further reinforces our hypothesis that a model with genuine semantic understanding should maintain consistent performance when transitioning between \emph{definition}-based and \emph{example}-based evaluations.

\paragraph{Validation against Spanish \wic{}.} Following the same procedure as in English, we evaluated the results obtained by \SB{} against the Spanish \wic{} dataset, as shown in \cref{fig:scatter_ES}. In this evaluation, we report only the results for the \SBdef{} variant, since the Spanish dictionary used in our setup does not provide example sentences for every sense. The results yield a promising Spearman's correlation coefficient of $\rho=0.765$, indicating that our method is also effective and reliable for moderately resourced languages such as Spanish. Furthermore, we confirm that \SB{} provides higher discriminative power than \wic{}, with results showing again a noticeably wider spread across models in Spanish.

\paragraph{Validation against Basque \wic{}.} Results for Basque are presented in \cref{fig:scatter_EU}. The Basque results exhibit a lower but still statistically significant rank correlation ($\rho=0.657$, p-value $<0.05$). This can be largely attributed to the fact that most models perform at near-random levels on the Basque \wic{} (as shown below in~\cref{ssec:results}), leading to almost random rankings. Although this phenomenon also affects \SB{} to some extent, its results appear more coherent: Basque-specialized models consistently outperform the others, indicating that \SB{} is still capable of capturing meaningful distinctions even under low-resource conditions. Interestingly, \SB{} remains more sensitive than \wic{} to relative performance differences, even when overall accuracy is low.

\subsection{Robustness analysis}
\label{ssec:robustness}

In addition to the validation with \wic{} on several languages, we also explored the robustness of our method regarding the total amount of examples in the benchmark required to produce a significantly well correlated ranking and the number of in-context examples (zero-shot vs few-shot). The following paragraphs discuss the results we obtained.

\paragraph{Impact of the number of instances.}
\cref{fig:boostraping_stability} shows the Spearman's correlation curves as a function of the number of test instances. To conduct this analysis, we performed bootstrapping over 100 iterations on subsets randomly sampled from the original test set of 1,000 instances, with sample sizes ranging from 50 to 1,000 instances. This procedure was repeated for the two \SB{} variants (\SBdef{} and \SBex{}) across four difficulty levels: \textit{random}, \textit{easy}, \textit{medium}, and \textit{hard}. For each point, we report the correlation along with the corresponding 95\% confidence interval.

Overall, all variants converge rapidly to a high correlation (above 0.9), with only marginal gains observed beyond 500 instances. As expected, confidence intervals narrow as the number of data points increases, indicating greater stability with larger sample sizes. Both variants, \emph{from definitions} and \emph{from examples}, display comparable correlation curves; however, contrary to our expectations, \SBdef{} yields consistently smaller confidence intervals. A plausible explanation for this behavior is that the examples generated by the models are more similar among them, leading to slightly more stable results for the \SBex{} variant.

\begin{figure}
    \includegraphics[width=\linewidth]{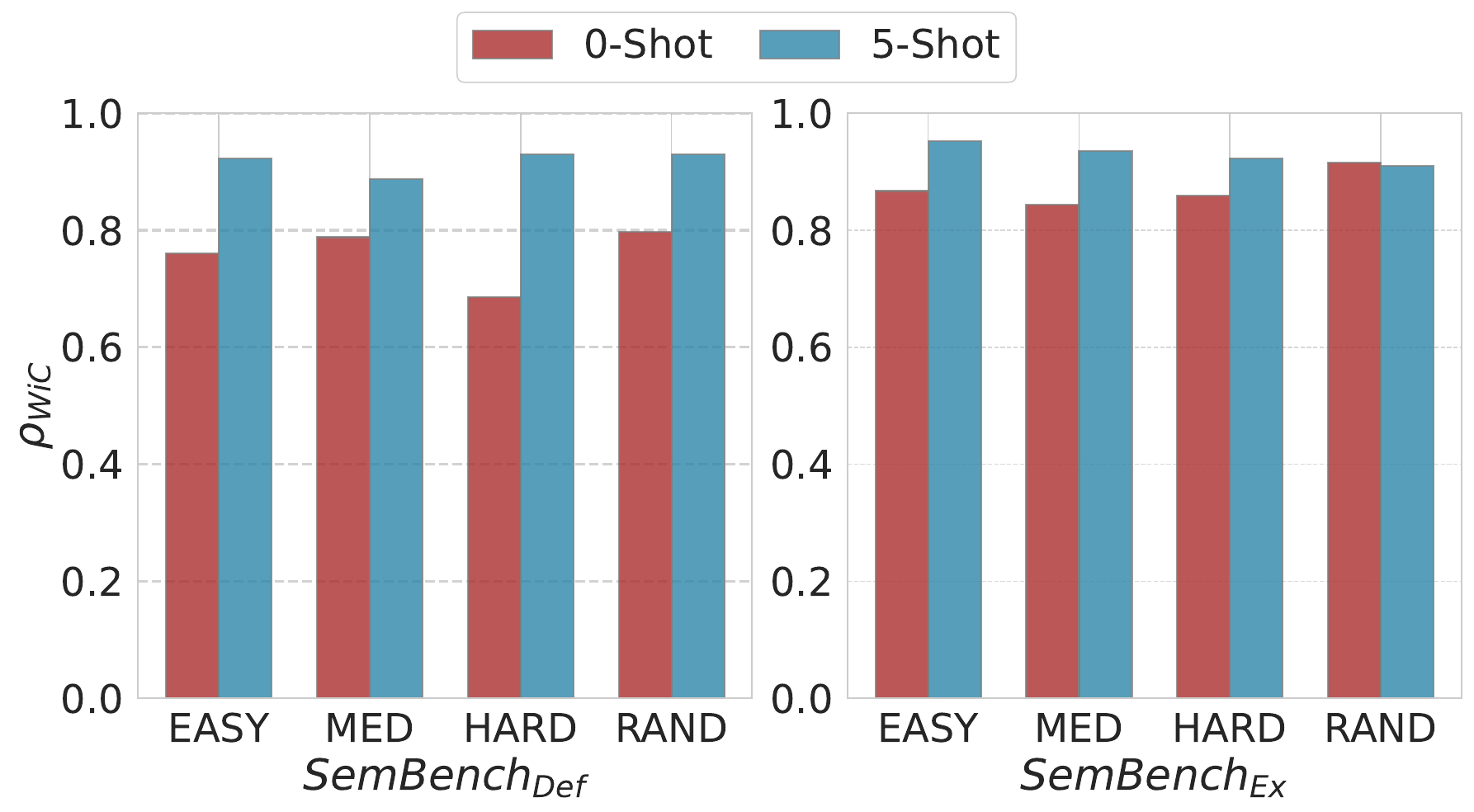}       
    \caption{Spearman's $\rho$ correlation comparison between \textcolor[HTML]{bb5858}{0-Shot} and \textcolor[HTML]{589ebc}{5-Shot} in \SBdef{} (left) and \SBex{} (right) against \wic.}
    \label{fig:bar_plot}
\end{figure}

\paragraph{Zero-Shot vs. Few-Shot Results Comparison.}
All results reported so far were obtained by prompting the LLMs with five in-context examples. This few-shot configuration helps guide the model toward the desired task format and can improve its alignment with the evaluation objective. However, this setup also introduces a small amount of manual effort, since examples must be created when they are not available in the dictionary. Although generating five examples per sense represents only minimal human intervention, it raises an important question: to what extent do these examples influence the overall performance and model ranking? To address this, we compare the results obtained under two settings---zero-shot (no examples provided) and few-shot (five examples)---and report the difference in Spearman’s $\rho$ between the two.

\cref{fig:bar_plot} summarizes these results across all experimental configurations. Overall, the correlations remain high in both scenarios, indicating that the models capture the intended task dynamics even without explicit examples. However, we observe decreases in correlation values when transitioning from few-shot to zero-shot prompting. The largest performance gap appears in the \SBdef{} approach, where the removal of examples leads to a more noticeable drop in correlation. This behavior is expected, as \SBdef{} involves additional intermediate steps in the evaluation pipeline. The model must first generate a usage example from the definition and then infer the definition again from that generated example, which introduces additional opportunities for variation. Consequently, the presence of in-context examples may provide useful guidance that stabilizes this multi-step generation process.

The correlation remains remarkably stable in \SBex{}, demonstrating that the approach maintains its reliability and consistency even in a zero-shot setting. Overall, these results underscores the potential of \SB{} as a scalable evaluation framework, minimizing the dependence on handcrafted examples while preserving strong agreement with human or reference-based judgments. In practical terms, these results suggest that \SB{} can be effectively applied in scenarios where annotated data or examples are scarce, further reinforcing its suitability for low-resource or multilingual contexts.

\subsection{\SB{} results}
\label{ssec:results}

\begin{table}
    \begin{tblr}{
        width=\linewidth,
        colspec={l|XXX|XXX},
        cells={c,font=\scriptsize},
        row{1,2}={font=\scriptsize\bfseries},
        column{1}={l},
        rowsep=0.65pt
    }
        \toprule
                       & \SetCell[c=3]{c} \SBex{} &&& \SetCell[c=3]{c} \SBdef{} \\
        \textbf{Model Name} &
          \textsc{easy} &
          \textsc{med} &
          \textsc{hard} &
          \textsc{easy} &
          \textsc{med} &
          \textsc{hard} \\
          \midrule
        Gemma 3$_{4B}$ & 78.10 & 67.20 & 61.40 & 79.30 & 68.80 & 63.70 \\
        Qwen3$_{4B}$ & \textbf{91.20} & \textbf{86.20} & \textbf{82.00} & \textbf{92.00} & \textbf{86.40} & \textbf{83.50} \\
        \midrule
        Llama 2$_{7B}$ & 78.70 & 68.50 & 66.00 & 75.50 & 64.60 & 60.80 \\
        Llama 3.1$_{8B}$ & 90.80 & 84.00 & 81.60 & 91.20 & 84.90 & 82.20 \\
        Latxa$_{8B}$ & 87.60 & 81.10 & 77.00 & 86.00 & 77.90 & 74.60 \\
        Qwen3$_{8B}$ & \textbf{93.10} & \textbf{87.80} & \textbf{86.10} & \textbf{95.00} & \textbf{90.10} & \textbf{86.70} \\
        \midrule
        Gemma 3$_{12B}$ & 88.40 & 83.50 & 78.10 & 90.30 & 85.00 & 79.60 \\
        Qwen3$_{14B}$ & \textbf{93.90} & \textbf{89.10} & \textbf{85.70} & \textbf{96.20} & \textbf{91.50} & \textbf{89.90} \\
        \midrule
        Gemma 3$_{27B}$ & 89.70 & 83.80 & 80.00 & 91.00 & 84.60 & 80.40 \\
        Qwen3$_{32B}$ & \underline{\textbf{95.30}} & \underline{\textbf{89.40}} & \underline{\textbf{86.70}} & \underline{\textbf{96.90}} & \underline{\textbf{92.90}} & \underline{\textbf{91.30}} \\
        \midrule
        Llama 3.1$_{70B}$ & \textbf{94.00} & \textbf{89.10} & 85.50 & 95.50 & 90.50 & \textbf{89.30} \\
        Latxa$_{70B}$ & 93.90 & 88.80 & \textbf{86.30} & \textbf{95.60} & \textbf{90.80} & 87.80 \\
        \bottomrule
    \end{tblr}
    \caption{\SB{} English results in 5-Shot scenario in the different difficulty levels. Bold-case indicates the best results by group. Underline indicates the best results overall.}
    \label{tab:EN_results}
\end{table}

\begin{table}
    \begin{tblr}{
        width=\linewidth,
        colspec={l|XX|XX|XX},
        cells={c,font=\scriptsize},
        row{1,2}={font=\scriptsize\bfseries},
        column{1}={l},
        rowsep=0.65pt
    }
        \toprule
                   & \SetCell[c=2]{c} English && \SetCell[c=2]{c} Spanish && \SetCell[c=2]{c} Basque \\
        Model Name & SB & \wic{} & SB & \wic{} & SB & \wic{} \\
        \midrule
        Gemma 3$_{4B}$ & 70.70 & 60.16 & 62.60 & 53.90 & 55.08 & 50.20 \\
        Qwen3$_{4B}$ & \textbf{88.80} & \textbf{65.16} & \textbf{78.50} & \textbf{57.60} & \textbf{59.82} & \textbf{52.00} \\
        \midrule
        Llama 2$_{7B}$ & 67.00 & 57.36 & 62.40 & 54.10 & 55.53 & 49.95 \\
        Llama 3.1$_{8B}$ & 87.90 & 65.80 & 72.70 & 56.50 & 59.82 & 52.60 \\
        Latxa$_{8B}$ & 81.80 & 62.51 & 69.70 & 55.40 & \textbf{63.43} & 51.60 \\
        Qwen3$_{8B}$ & \textbf{90.90} & \textbf{68.25} & \textbf{80.80} & \textbf{60.40} & 58.47 & \textbf{54.40} \\
        \midrule
        Gemma 3$_{12B}$ & 86.00 & 64.77 & 72.30 & 59.50 & \textbf{63.21} & 55.55 \\
        Qwen3$_{14B}$ & \textbf{93.20} & \textbf{68.11} & \textbf{81.50} & \textbf{60.20} & \textbf{63.21} & \textbf{55.90} \\
        \midrule
        Gemma 3$_{27B}$ & 86.90 & 66.83 & 73.60 & 58.40 & 58.47 & 55.85 \\
        Qwen3$_{32B}$ & \underline{\textbf{93.90}} & \underline{\textbf{70.61}} & \textbf{80.30} & \underline{\textbf{61.10}} & \textbf{64.56} & \underline{\textbf{57.05}} \\
        \midrule
        Llama 3.1$_{70B}$ & \textbf{93.10} & \textbf{68.69} & 78.40 & \textbf{60.40} & 65.91 & 55.65 \\
        Latxa$_{70B}$ & 92.70 & 68.65 & \underline{\textbf{80.90}} & 59.50 & \underline{\textbf{73.59}} & \textbf{56.65} \\
        \bottomrule
    \end{tblr}
    \caption{Multilingual results in \wic{} and \SBex{} (SB) with \textsc{rand} strategy at 5-shot scenario. Bold-case indicates the best results by group. Underline indicates the best results overall.}
    \label{tab:multilingual_results}
\end{table}

Overall \SB{} performance of all evaluated models across languages and difficulty levels is summarized in~\cref{tab:EN_results,tab:multilingual_results}. The former presents English results for both \SBdef{} and \SBex{} variants by difficulty level, while the latter extends the comparison to the multilingual setup (English, Spanish, and Basque) using the random difficulty strategy.

\paragraph{Impact of \SB{} difficulty.} We observe a consistent trend across all configurations: as task difficulty increases, model performance systematically decreases, following the pattern \textsc{easy} $>$ \textsc{med} $>$ \textsc{hard}. This confirms that \SB{}'s difficulty control mechanism effectively captures the semantic complexity of the generated instances, making the use of difficulty-controlled subsets convenient when available (although random sampling also yields stable results). Similarly, both \SB{} variants yield closely aligned results, with \SBdef{} generally producing slightly higher scores than \SBex{}, suggesting that examples generated by LLMs provide richer contextual cues that can lead to more accurate sense discrimination when generating definitions.

\paragraph{Impact of model family and size.} When examining specific model families, the Qwen3 series stands out as the best overall performer, surpassing even larger models in most configurations. This advantage likely stems from its enhanced reasoning-oriented training, which appears to contribute more precise sense discrimination and definition generation. Larger variants within each family (e.g., Llama 3.1 70B and Qwen3 32B) consistently outperform their smaller counterparts, capturing the positive effect of scaling on semantic generalization. Conversely, smaller models such as Gemma 4B and Llama 7B show pronounced drops in accuracy as difficulty increases. Interestingly, linguistically specialized models (Latxa 8B and 70B) demonstrate that language adaptation can partially offset size limitations, especially in low-resource settings.

\paragraph{Impact of \SB{} language.} The multilingual results in~\cref{tab:multilingual_results} provide further insight into performance trends across resource availability levels. As expected, absolute performance decreases from English to Spanish and further to Basque, mirroring the availability of linguistic resources and, potentially, the quality of underlying dictionaries. Still, \SB{} successfully preserves meaningful model rankings even in low-resource settings, with Basque-specialized models (namely, Latxa) outperforming general-purpose ones in the Basque test---unlike in \wic{}. That is, \SB{} can capture language-specific semantic competence effectively, even when absolute accuracy is limited.

\section{Conclusions}
\label{sec:conclusions}

We introduce \SB{}, a framework for evaluating the semantic competence of Large Language Models (LLMs) through controlled text generation from dictionary definitions. By relying solely on sense definitions and a sentence encoder, \SB{} provides a fully automatic, language-independent approach that eliminates the need for manually annotated data.

Our experiments confirm that \SB{} produces model rankings strongly aligned with those obtained on standard \wic{} benchmarks, validating it as an effective and faithful alternative to semantic evaluation. At the same time, the wider range of results observed in \SB{} indicated a higher discriminative capacity, enabling to separate model performances more clearly and reveal subtler differences in semantic competence. Multilingual evaluations further demonstrate that \SB{} maintains consistency across English, Spanish, and Basque. Particularly in Basque, domain-adapted models outperform general-purpose ones, unlike in \wic{}, which shows that \SB{} can better capture language-specific semantic competence even under limited-resource conditions.

Ablation studies also confirm the scalability and efficiency of the approach: as few as 250 instances are sufficient to produce interpretable results, and only marginal gains are observed beyond 500. In addition, the proposed difficulty-control heuristic accurately reflects task complexity, with performance decreasing predictably from \textsc{easy} to \textsc{hard} sets while preserving strong correlation with \wic{}.

Overall, \SB{} offers a consistent, interpretable, and language-independent methodology for assessing semantic competence in LLMs. Its robustness across configurations, model sizes, and languages confirm its value as a lightweight and convenient alternative to traditional manually annotated benchmarks. Most importantly, by depending only on dictionary definitions and general-purpose encoders, \SB{} can be readily applied to new and under-resourced languages where benchmarks such as \wic{} or other annotated datasets are not available.

\section*{Limitations}
\label{sec:limitations}
While \SB{} shows strong alignment with traditional semantic benchmarks, several limitations remain. First, \SB{} relies on a strong multilingual encoder (i.e., EmbeddingGemma 300M), which has demonstrated competitive performance. However, our dependence on a single encoder may introduce biases and may fail to distinguish between subtle senses. Exploring alternative or ensemble encoders could further enhance robustness. Second, our evaluation focused on open-weight models of different sizes. However, including commercial LLMs such as Claude or GPT would help assess whether \SB{} can effectively rank state-of-the-art language models. Finally, it would be interesting to compare \SB{} results with more general evaluation efforts, such as LLMArena, to examine whether the semantic competence measured here aligns with broader model preferences and overall perceived quality.

\section*{Acknowledgments}
This work has been partially supported by the European Union under Horizon Europe (Project LUMINOUS, grant number 101135724) and the Spanish Ministry of Science, Innovation, and Universities (Project HumanAIze, grant number AIA2025-163322-C61). It was also funded by the Basque Government (IKER-GAITU project) and the Ministerio para la Transformación Digital y de la Función Pública - Funded by EU – NextGenerationEU within the framework of the project Desarrollo de Modelos ALIA. Mikel Zubillaga holds a PhD grant from the University of the Basque Country UPV/EHU (PIF24/04).


\bibliographystyle{lrec2026-natbib}
\bibliography{lrec2026-example}

\appendix

\section{Prompts}\label{app:prompts}
In order to define the tasks of SemBench for LLMs, we designed two complementary prompt configurations: \textit{i)} generating an example from a definition, and \textit{ii)} generating a definition from an example. In line with best practices for prompting modern LLMs, each configuration consists of a \textit{System Prompt} and a \textit{User Prompt}.
\subsection{Generating an example from a definition}
\paragraph{System Prompt:} You are an expert \textit{\{language\}} lexicographer. Your task is to generate ONLY ONE example in \textit{\{language\}} of the usage of a word, given a definition of that word. Please, provide JUST the example---DO NOT include the definition or any other further explanation.
\paragraph{User Prompt:} Given the \textit{\{part-of-speech\}} `\textit{\{word\}}' and its sense in this definition: `\textit{\{definition\}}', generate one usage example of the word for that sense. Give JUST the example without further explanation.
\subsection{Generating a definition from an example}
\paragraph{System Prompt:} You are an expert \textit{\{language\}} lexicographer. Your task is to generate a dictionary definition in \textit{\{language\}} of a word, given some example sentences of the word. Please, provide JUST the definition---DO NOT include the example or any other further explanations.
\paragraph{User Prompt:} Given the \textit{\{part-of-speech\}} `\textit{\{word\}}' and its sense in this example: `\textit{\{example\}}', generate the definition of the word for that sense. Give JUST the definition without further explanation.

\end{document}